%% file: 3134.tex
\documentclass[runningheads]{llncs}
\usepackage{graphicx}
\usepackage{amsmath,amssymb} %
\usepackage{color}
\usepackage[width=122mm,left=12mm,paperwidth=146mm,height=193mm,top=12mm,paperheight=217mm]{geometry}

\usepackage{wrapfig}
\usepackage{todo}
\usepackage{caption}

\input{ce_common}

\usepackage[colorlinks=true]{hyperref}

\begin{document}
\pagestyle{headings}
\mainmatter

\title{Learning SO(3) Equivariant Representations with Spherical CNNs} %

\titlerunning{Learning SO(3) Equivariant Representations with Spherical CNNs}

\authorrunning{C. Esteves, C. Allen-Blanchette, A. Makadia and K. Daniilidis}

\author{\small Carlos Esteves\textsuperscript{1}, Christine Allen-Blanchette\textsuperscript{1}, Ameesh Makadia\textsuperscript{2}, Kostas Daniilidis\textsuperscript{1}}

\institute{\small \textsuperscript{1}GRASP Laboratory, University of Pennsylvania \quad \textsuperscript{2}Google
{\small \{machc,allec,kostas\}@seas.upenn.edu} \quad makadia@google.com}

\maketitle

\begin{abstract}
We address the problem of 3D rotation equivariance in convolutional
neural networks. 3D rotations have been a challenging nuisance in 3D
classification tasks requiring higher capacity and extended data
augmentation in order to tackle it. We model 3D data with
multi-valued spherical functions and we propose a novel spherical
convolutional network that implements exact convolutions on the sphere
by realizing them in the spherical harmonic domain. Resulting filters
have local symmetry and are localized by enforcing smooth spectra. We
apply a novel pooling on the spectral domain and our operations are
independent of the underlying spherical resolution throughout the
network. We show that networks with much lower capacity and without
requiring data augmentation can exhibit performance comparable to the
state of the art in standard retrieval and classification benchmarks.

\end{abstract}
\blfootnote{\url{http://github.com/daniilidis-group/spherical-cnn}}

\input{intro}
\input{related}

\input{math}

\input{method}
\input{experiments}

\section{Conclusion}

We presented Spherical CNNs, which leverage spherical convolutions to achieve equivariance to \SO(3) perturbations.
The network is applied to 3D object classification, retrieval, and alignment,
but has potential applications in spherical images such as panoramas,
or any data that can be represented as a spherical function.
We show that our model can naturally handle arbitrary input orientations, requiring relatively few parameters and small input sizes.

\subsubsection{Acknowledgments:}
We are grateful for support through the following grants: NSF-DGE-0966142 (IGERT), NSF-IIP-1439681 (I/UCRC), NSF-IIS-1426840, NSF-IIS-1703319, NSF MRI 1626008, ARL RCTA W911NF-10-2-0016, ONR N00014-17-1-2093, and by Honda Research Institute.

\newpage
\bibliographystyle{splncs}
\bibliography{refs.bib}

\end{document}

%% file: ce_common.tex
\usepackage{mathtools}

\newcommand{\norm}[1]{\left\lVert#1\right\rVert}

\newcommand{\SO}{\ensuremath{\mathbf{SO}}}

\newcommand{\etal}{\textit{et al}. }

\newcommand\blfootnote[1]{%
  \begingroup%
  \renewcommand\thefootnote{}\footnote{#1}%
  \addtocounter{footnote}{-1}%
  \endgroup%
}

%% file: intro.tex
\section{Introduction}

One of the reasons for the tremendous success of convolutional neural networks (CNNs) is their equivariance to translations in euclidean spaces and the resulting invariance to local deformations. Invariance with respect to other nuisances has been traditionally addressed with data augmentation while non-euclidean inputs like point-clouds have been approximated by euclidean representations like voxel spaces.
Only recently, equivariance has been addressed with respect to other groups \cite{cohen2016group,worrall2017harmonic} and CNNs have been proposed for manifolds or graphs \cite{bruna2013learning,bronstein2017geometric,s.2018spherical}.

Equivariant networks retain information about group actions on the input and on the feature maps throughout the layers of a network. Because of their special structure, feature transformations are directly related to spatial transformations of the input. Such equivariant structures yield a lower network capacity in terms of unknowns than alternatives like the Spatial Transformer \cite{jaderberg2015spatial} where a canonical transformation is learnt and applied to the original input. 

\begin{figure}
  \centering
  \begin{minipage}{0.45\textwidth}
    \includegraphics[width=\textwidth]{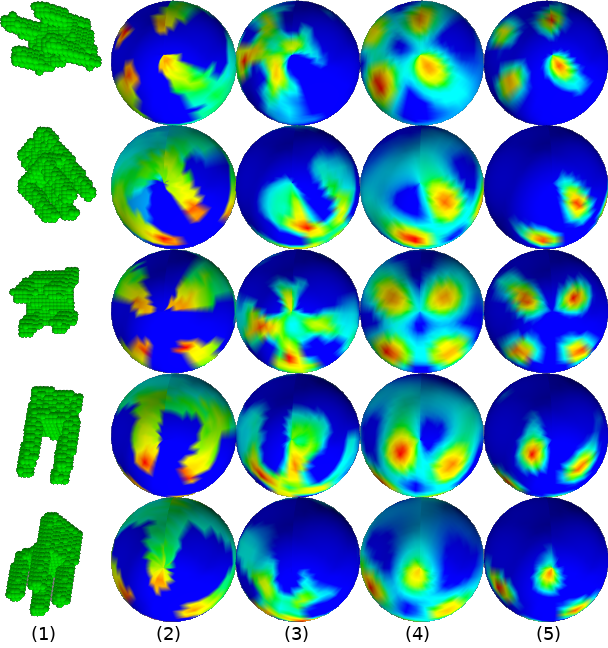}
    \caption{\small Columns: (1) input, (2) initial spherical representation, (3-5) learned feature maps.  Activations of chair legs illustrate rotation equivariance.}
    \label{fig:intro}
  \end{minipage}
  \hfill
  \begin{minipage}{0.5\textwidth}
    \includegraphics[width=\textwidth]{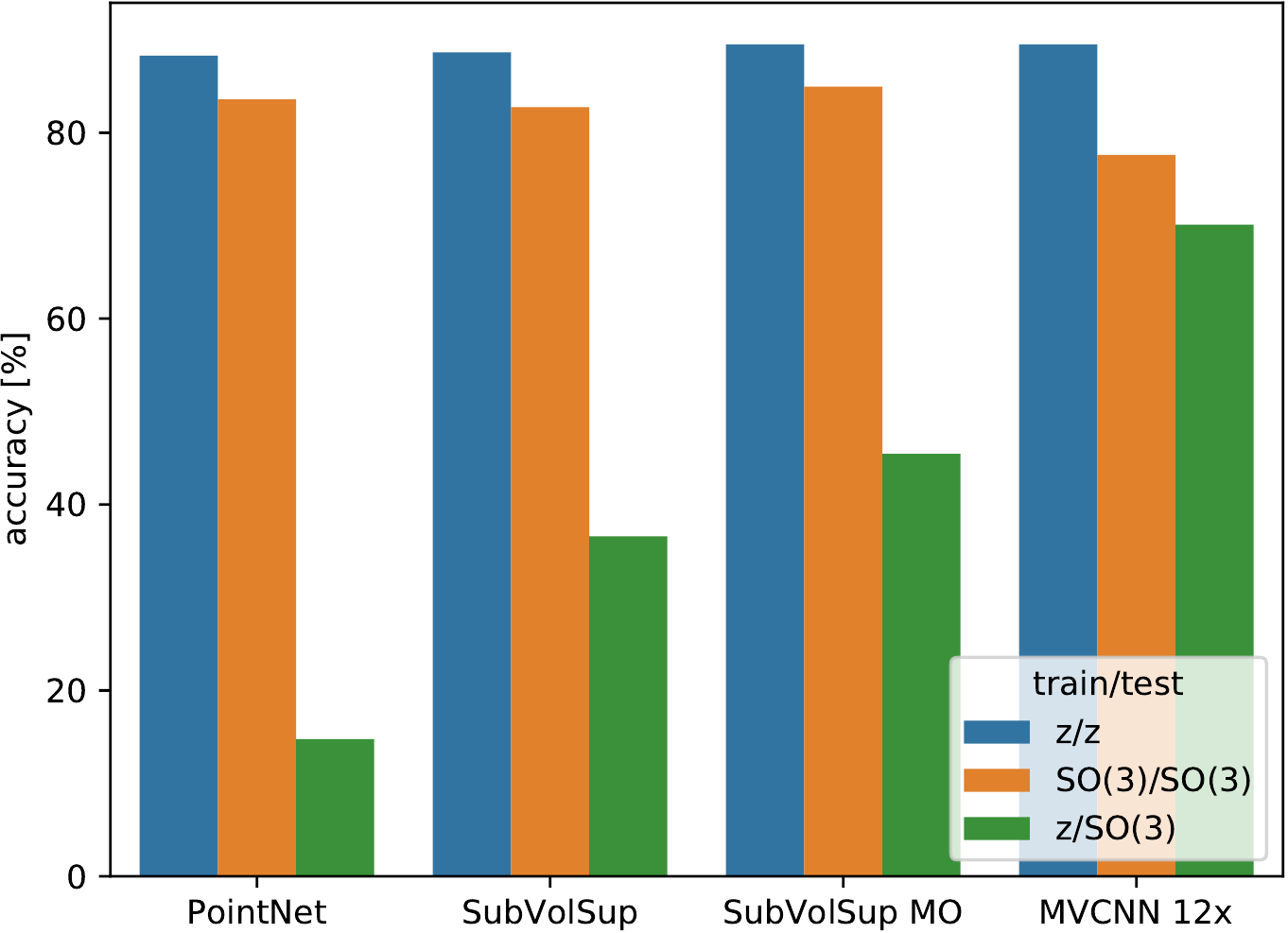}
    \caption{\small ModelNet40 classification for 
  point cloud~\cite{qi2017pointnet}, volumetric~\cite{Qi2016volumetric}, and multi-view~\cite{su2015multi} methods. The significant drop in accuracy illustrates that conventional methods do not generalize to arbitrary (SO(3)/SO(3)) and unseen orientations (z/SO(3)).}
  \label{fig:methods-zz-so3so3-zso3}
  \end{minipage}
\end{figure}

In this paper, we are primarily interested in analyzing 3D data for alignment, retrieval or classification. Volumetric and point cloud representations have yielded translation and scale invariant approaches: Normalization of translation and scale can be achieved by setting the object's origin to its center and constraining its extent to a fixed constant. However, 3D rotations remain a challenge to current approaches (Figure~\ref{fig:methods-zz-so3so3-zso3} illustrates how classification performance for conventional methods suffers when arbitrary rotations are introduced).

In this paper, we model 3D-data with  spherical functions valued in ${\mathbb{R}}^n$ and introduce a novel equivariant convolutional neural network with spherical inputs (Figure~\ref{fig:intro} illustrates the equivariance). We clarify the difference between convolution that has spherical outputs and correlation that has outputs in the rotation group $\mathbf{SO}(3)$ and we apply exact convolutions that yield zonal filters, i.e. filters with constant values along the same latitude. Convolutions cannot be applied with spatially-invariant impulse responses (masks), but can be exactly computed in the spherical harmonic domain through pointwise multiplication. To obtain localized filters, we enforce a smooth spectrum by learning weights only on few anchor frequencies and interpolating between them, yielding, as additional advantage, a number of weights independent of the spatial resolution.

It is natural then to apply pooling in the spectral domain.
Spectral pooling has the advantage that it retains equivariance while spatial pooling on the sphere is only approximately equivariant.
We also propose a weighted averaging pooling where the weights are proportional to the cell area.
The only reason to return to the spatial domain is the rectifying nonlinearity, which is a pointwise operator.

We perform 3D retrieval, classification, and alignment experiments. Our aim is to show that we can achieve near state of the art performance with a much lower network capacity, which we achieve for the SHREC'17 \cite{savva2017shrec} contest and ModelNet40 \cite{Wu20153Dshapenets} datasets.

Our main contributions can be summarized as follows: 
\begin{itemize}
\setlength\itemsep{0em}
\item We propose the first neural network based on spherical convolutions.
\item We introduce pooling and parameterization of filters in the spectral domain, with enforced spatial localization and capacity independent of the resolution.
\item Our network has much lower capacity than non-spherical networks applied on 3D data without sacrificing  performance.
\end{itemize}

We start with the related work, then introduce the mathematics of group and in particular sphere convolutions, and details of our network. Last, we perform extensive experiments on retrieval, classification, and alignment.

%% file: related.tex
\section{Related work}
\label{sec:related_work}
We will start describing related work on group equivariance, in particular equivariance on the sphere, then delve into CNN representations for 3D data.

Methods for enabling equivariance in CNNs can be divided in two groups. In the first, equivariance is obtained by constraining filter structure similarly to Lie generator
based approaches \cite{segman1992canonical,hel1996canonical}. Worral \etal \cite{worrall16_harmon_networ} use filters derived from the complex harmonics achieving both rotational and translational equivariance.
The second group requires the use of a filter orbit which is itself equivariant to obtain group equivariance. Cohen and Welling \cite{cohen2016group} convolve with the orbit of a learned filter and prove the equivariance of
group-convolutions and
preservation of rotational equivariance in the presence of rectification and pooling.
Dieleman \etal \cite{dieleman2015rotation} process elements of the image orbit individually and use
the set of outputs for classification.
Gens and Domingos \cite{gens2014deep} produce maps of finite-multiparameter groups,
Zhou \etal \cite{Zhou_2017_CVPR} and Marcos \etal \cite{marcos16_rotat_equiv_vector_field_networ} use a rotational filter orbit to produce oriented feature maps and rotationally invariant features, and
Lenc and Vedaldi \cite{lenc2015understanding} propose a transformation layer which acts as a group-convolution
by first permuting then transforming by a linear filter.

Recently, a body of work on Graph Convolutional Networks (GCN) has emerged.
There are two threads within this space, spectral \cite{bruna2013spectral,defferrard2016convolutional,kipf2016semi} and spatial \cite{boscaini2016learning,masci2015geodesic,monti2016geometric}.
These approaches learn filters on irregular but structured graph representations.
These methods differ from ours in that we are looking to explicitly learn equivariant
and invariant representations for 3D-data modeled as spherical functions under rotation.
While such properties are difficult to construct for general manifolds, we leverage the
group action of rotations on the sphere.

Most similar to our approach and developed in parallel\footnote{the first version of this work was submitted to CVPR on 11/15/2017, shortly after we became aware of Cohen \etal \cite{s.2018spherical}  ICLR submission on 10/27/2017.} is \cite{s.2018spherical},
which uses spherical correlation to map spherical inputs to features on \SO(3),
then processed with a series of convolutions on \SO(3).
The main difference is that we use spherical convolutions, which are potentially one order of magnitude faster, with smaller (one fewer dimension) filters and feature maps.
In addition, we enforce smoothness in the spectral domain that results in better localization of the receptive fields on the sphere and we perform pooling in two different ways,
either as a low-pass in the spectral domain or as a weighted averaging in the spatial domain.
Moreover, our method outperforms \cite{s.2018spherical} in the SHREC'17 benchmark.

Spherical representations for 3D-data are not novel and have been used for retrieval tasks before the deep learning era \cite{frome2004recognizing,kazhdan2002harmonic} because of their invariance properties and efficient implementation of spherical correlation \cite{makadia2010spherical}. In 3D deep learning, 
 the most natural adaptation of 2D methods was to use a
voxel-grid representation of the 3D object and amend the 2D CNN framework to use
collections of 3D filters for cascaded processing in the place of conventional 2D filters.
Such approaches require a tremendous amount of computation to achieve very basic voxel resolution and need a much higher capacity.

Several attempts have been made to use CNNs to 
produce discriminative representations from volumetric data.
3D ShapeNets \cite{Wu20153Dshapenets} and VoxNet \cite{daniel2015voxnet} propose a fully-volumetric network with 3D convolutional layers followed by fully-connected layers.
Qi \etal \cite{Qi2016volumetric} observe significant overfitting
when attempting to train the aforementioned
end-to-end and choose to amend the technique using
subvolume classification as an auxiliary task, and also propose an alternate 3D CNN which
learns to project the volumetric representation to a 2D representation, then processed using a conventional 2D CNN architecture.
Even with these adaptations, Qi \etal \cite{Qi2016volumetric} are challenged by
overfitting and suggest augmentation in the form of orientation pooling
as a remedy.
Qi \etal \cite{qi2017pointnet} also present an attempt to train a neural network that operates directly on point clouds.
Currently, the most successful approaches are view-based, operating in rendered views of the 3D object \cite{su2015multi,Qi2016volumetric,kanezaki16_rotat,bai2016gift}.
The high performance of these methods is in part due to the use of large pre-trained 2D CNNs (on ImageNet, for instance).

%% file: math.tex
\section{Preliminaries}
\label{sec:math}

\subsection{Group Convolution}
Consideration of symmetries, in particular rotational symmetries, naturally evokes notions
of the Fourier Transform. In the context of deriving rotationally invariant representations, the
Fourier Transform is particularly appealing since it exhibits invariance to rotational deformations
up to phase (a truly invariant representation can be achieved through application of the modulus
operator).

To leverage this property for 3D shape analysis, it is necessary to construct a rotationally equivariant representation of our 3D input.
For a group $G$ and function $f:E\rightarrow F$, $f$ is said to
be equivariant to transformations $g\in G$ when
\begin{equation}
f(g\circ x) = g'\circ f(x), \quad x\in E
\end{equation}
where $g$ acts on elements of $E$ and $g'$ is the corresponding group action which transforms
elements of $F$. If $E=F$, $g=g'$. A straightforward example of an equivariant representation is
an orbit. For an object $x$, its orbit $O(x)$ with respect to the group $G$ is defined
\begin{equation}
  O(x) = \{ g\circ x\; |\; \forall g\in G\}.
\end{equation}
Through this example it is possible to develop an intuition into the equivariance of the group
convolution; convolution can be viewed as the inner-products of some function $f$ with all
elements of the orbit of a ``flipped'' filter $h$. Formally, the group convolution is defined as
\begin{equation}
(f \star_G h)(x) = \int_{g \in G} f(g \circ \eta) h(g^{-1} \circ x) \, dg,
\end{equation}
where $\eta$ is typically a canonical element in the domain of $f$ (e.g. the origin if $E = \mathbb{R}^n$, or $I_n$ if $E = \SO(n)$).
The familiar convolution on the plane is a special case of the group convolution
with the group $G=\mathbb{R}^2$ with addition,
\begin{equation}
  \begin{aligned}
      (f \star h)(x) = \int_{g \in \mathbb{R}^2} f(g \circ \eta) h(g^{-1} \circ x) \, dg = \int_{g \in \mathbb{R}^2} f(g) h(x-g) \, dg.
  \end{aligned}
\end{equation}
The group convolution can be shown to be equivariant. For any $\alpha \in G$,
\begin{equation}
  \begin{aligned}
      ((\alpha^{-1} \circ f)\star_{G} h)(x) = (\alpha^{-1} \circ (f\star_G h))(x).
  \end{aligned}
\end{equation}

\subsection{Spherical harmonics}
\label{sec:spher-conv}
Following directly the preliminaries above, we can define convolution of spherical signal $f$ by a spherical filter $h$ with respect to the group of 3D rotations $\mathbf{SO}(3)$:
\begin{equation}
  \begin{aligned}
      (f \star_G h)(x) = \int_{g \in \mathbf{SO}(3)} f(g \eta) h(g^{-1} x) \, dg, \label{eq:sphconveq}
  \end{aligned}
\end{equation}
where $\eta$ is north pole on the sphere.

To implement \eqref{eq:sphconveq}, it is desirable to sample the sphere with well-distributed and compact cells with transitivity (rotations exist which bring cells into coincidence).  Unfortunately, such a discretization does not exist \cite{thurston97geotop}. Neither the familiar sampling by latitude and longitude nor the uniformly distributed sampling according to Platonic solids satisfies all constraints. These issues are compounded with the eventual goal of performing cascaded convolutions on the sphere.

To circumvent these issues, we choose to evaluate the spherical convolution in the spectral domain. This is possible as the machinery of Fourier analysis has extended the well-known convolution theorem to functions on the sphere: the Spherical Fourier transform of a convolution is the pointwise product of Spherical Fourier transforms (see \cite{arfken1966mathematical,driscoll1994computing} for further details). The Fourier transform and its inverse are defined on the sphere as follows~\cite{arfken1966mathematical}:

{\centering
\noindent\begin{minipage}{.5\linewidth}
  \begin{equation}
    f = \sum_{0 \le \ell \le b}\sum_{|m| \le \ell}\hat{f}_m^{\ell}Y_m^{\ell} \label{eq:isft},
  \end{equation}
\end{minipage}%
\begin{minipage}{.5\linewidth}
  \begin{equation}
    \hat{f}_m^{\ell} = \int_{S^2} f(x) \overline{Y_m^{\ell}} dx \label{eq:sft},
  \end{equation}
\end{minipage}}

\noindent where $b$ is the bandwidth of $f$, and $Y_m^{\ell}$ are the spherical harmonics of degree $\ell$ and order $m$.
We refer to (\ref{eq:sft}) as the Spherical Fourier Transform (SFT), and to (\ref{eq:isft}) as its inverse (ISFT).
Revisiting \eqref{eq:sphconveq}, letting $y = (f \star_G h)(x)$, the spherical convolution theorem~\cite{driscoll1994computing} gives us
\begin{equation}
  \hat{y}_m^{\ell} = 2\pi \sqrt{\\\frac{4\pi}{2\ell+1}} \hat{f}_m^{\ell} \hat{h}_0^{\ell} \label{eq:sphconv},
\end{equation}
To compute the convolution of a signal $f$ with a filter $h$, we first expand $f$ and $h$ into their spherical harmonic basis \eqref{eq:sft}, second compute the pointwise product \eqref{eq:sphconv}, and finally invert the spherical harmonic expansion \eqref{eq:isft}.

It is important to note that this definition of spherical convolution is unique from spherical correlation which produces an output response on \SO(3). Convolution here can be seen as marginalizing the angle responsible for rotating the filter about its north pole, or equivalently considering zonal filters on the sphere.

\subsection{Practical considerations and optimizations}
To evaluate the SFT, we use equiangular samples on the sphere according to the sampling theorem of~\cite{driscoll1994computing}
\begin{align}
  \hat{f}_m^{\ell} &= \frac{\sqrt{2\pi}}{2b}\sum_{j=0}^{2b-1}\sum_{k=0}^{2b-1} a_j^{(b)} f(\theta_j, \phi_k)\overline{Y_m^{\ell}}(\theta_j, \phi_k), \label{eq:dsft} \end{align}
where $\theta_j=\pi j/2b$ and $\phi_k=\pi k/b$ form the sampling grid, and $a_j^{(b)}$ are the sample weights.
Note that all the required operations are matrix pointwise multiplications and sums, which are differentiable and readily available in most automatic differentiation frameworks.
In our direct implementation, we precompute all needed $Y_m^{\ell}$, which are stored as constants in the computational graph.

\subsubsection{Separation of variables:} We also implement a potentially faster SFT based on separation of variables as shown in \cite{driscoll1994computing}.
Expanding $Y_m^{\ell}$ in (\ref{eq:dsft}), we obtain
\begin{equation}
  \begin{aligned}
  \hat{f}_m^{\ell} &= \sum_{j=0}^{2b-1}\sum_{k=0}^{2b-1} a_j^{(b)} f(\theta_j, \phi_k) q_m^{\ell} P_m^{\ell}(\cos{\theta_j})e^{-im\phi_k}   \\
  &= q_m^{\ell} \sum_{j=0}^{2b-1}a_j^{(b)} P_m^{\ell}(\cos{\theta_j}) \sum_{k=0}^{2b-1}f(\theta_j, \phi_k) e^{-im\phi_k},
  \end{aligned}
\end{equation}

\noindent where $P_m^{\ell}$ is the associated Legendre polynomial, and $q_m^{\ell}$ a normalization factor.
The inner sum can be computed using a row-wise Fast Fourier Transform
and what remains is an associated Legendre transform, which we compute directly.
The same idea is done for the ISFT.
We found that this method is faster when $b \ge 32$.
There are faster algorithms available \cite{driscoll1994computing,healy2003ffts}, which we did not attempt.

\subsubsection{Leveraging symmetry:} For real-valued inputs, $\hat{f}_{-m}^{\ell} = (-1)^{m}\overline{\hat{f}_{m}^{\ell}}$ (this follows from $\overline{Y_{-m}^{\ell}} = (-1)^m Y_m^{\ell}$). We thus need only compute half the coefficients ($m > 0$). Furthermore, we can rewrite the SFT and ISFT to avoid expensive complex number support or multiplication:
\begin{equation}
f = \sum_{0 \le \ell \le b} \left(\hat{f}_0^{\ell}Y_0^{\ell} + \sum_{m=1}^{\ell}  2\,\text{Re}(\hat{f}_m^{\ell})\text{Re}(Y_m^{\ell}) - 2\,\text{Im}(\hat{f}_m^{\ell})\text{Im}(Y_m^{\ell})\right).
\end{equation}

%% file: method.tex
\section{Method}
\label{sec:method}

\begin{figure}[t]
\centering
\includegraphics[width=\linewidth]{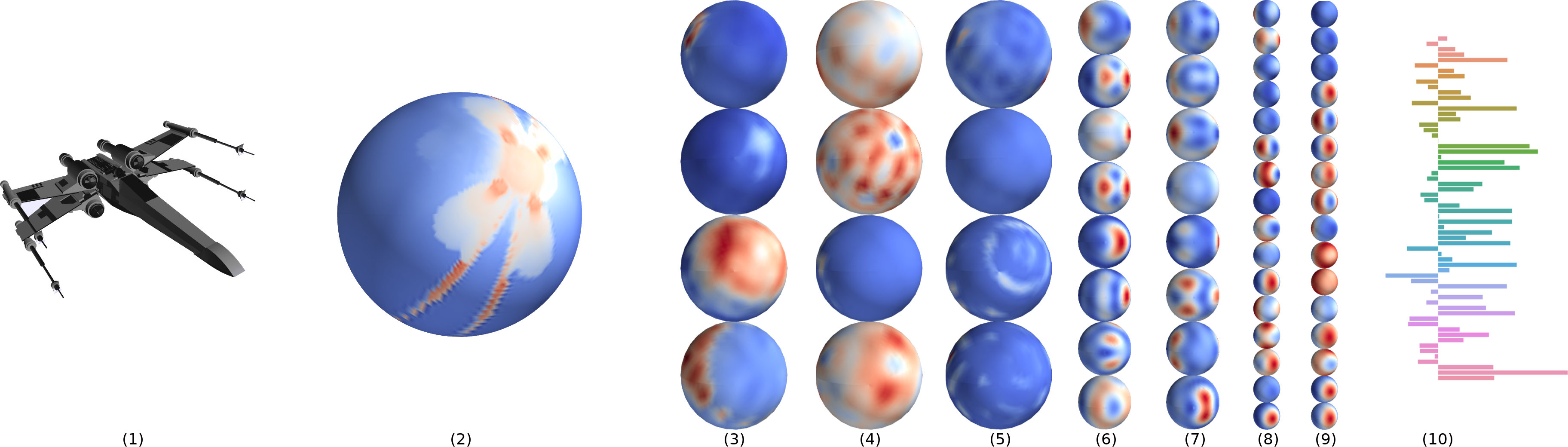}
\caption{
  Overview of our method.
  From left to right: a 3D model (1) is mapped to a spherical function (2),
  which passes through a sequence of spherical convolutions, nonlinearities and pooling,
  resulting in equivariant feature maps (3--9).
  We show only a few channels per layer.
  A global weighted average pooling of the last feature map results in a descriptor invariant to rotation (10),
  which can be used for classification or retrieval.
  The input spherical function (2) may have multiple channels, in this picture we show the distance to intersection representation.
  }
\label{fig:overview}
\end{figure}

Figure~\ref{fig:overview} shows an overview of our method.
We define a block as one spherical convolutional layer, followed by optional pooling, and  nonlinearity.
A weighted global average pooling is applied at the last layer to obtain an invariant descriptor.
This section details the architectural design choices.

\subsection{Spectral filtering}

In this section, we define the filter parameterization.
One possible approach would be to define a compact support around one of the poles and learn the values for each discrete location, setting the rest to zero.
The downside of this approach is that there are no guarantees that the filter will be bandlimited.
If it is not, the SFT will be implicitly bandlimiting the signal, which causes a discrepancy between the parameters and the actual realization of the filters.

To avoid this problem, we parameterize the filters in the spectral domain.
In order to compute the convolution of a function $f$ and a filter $h$, only the SFT coefficients of order $m=0$ of $h$ are used.
In the spatial domain, this implies that for any $h$, there is always a zonal filter (constant value per latitude) $h_z$, such that $\forall y,\, y * h = y * h_z$.
Thus, it only makes sense to learn zonal filters.

The spectral parameterization is also faster because it eliminates the need to compute the filter SFT, since the filters are defined in the spectral domain, which is the same domain where the convolution computed.

\subsubsection{Non-localized filters:}
A first approach is to parameterize the filters by all SFT coefficients of order $m=0$.
For example, given $32 \times 32$ inputs, the maximum bandwidth is $b=16$, so there are $16$ parameters to be learned ($\hat{h}_0^0, \ldots \hat{h}_0^{15} $). A downside is that the filters may not be local; however, locality may be learned.

\subsubsection{Localized filters:}
From Parseval's theorem and the derivative rule from Fourier analysis we can show that spectral smoothness corresponds to spatial decay.
This is used in the construction of graph-based neural networks \cite{bruna13_spect_networ_local_connec_networ_graph},
and also applies to the filters spanned by the family of spherical harmonics of order zero ($m=0$).

\begin{figure}[t]
\centering
\includegraphics[width=\linewidth]{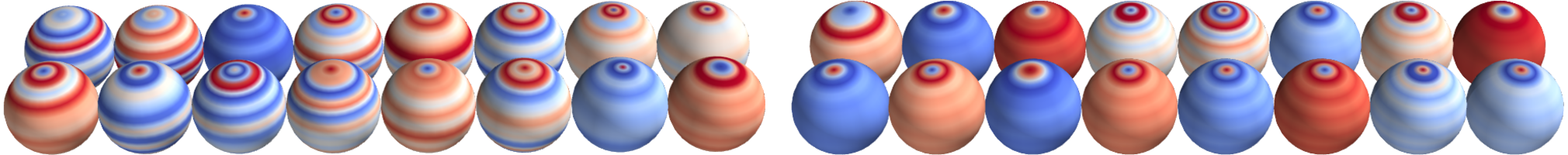}
\caption{
  Filters learned in the first layer.
  The filters are zonal.
  \emph{Left:} 16 nonlocalized filters. \emph{Right:} 16 localized filters.
  Nonlocalized filters are parameterized by all spectral coefficients (16, in the example).
  Even though locality is not enforced, some filters learn to respond locally.
  Localized filters are parameterized by a few points of the spectrum (4, in the example), the rest of the spectrum is obtained by interpolation.
  }
  \label{fig:conv0}
\end{figure}

To obtain localized filters, we parameterize the spectrum with anchor points.
We fix $n$ uniformly spaced degrees $\ell_i$ and learn the correspondent coefficients $f_0^{\ell_i}$.
The coefficients for the missing degrees are then obtained by linear interpolation, which enforces smoothness.
A second advantage is that the number of parameters per filter is independent of the input resolution.
Figure~\ref{fig:conv0} shows some filters learned by our model; the right side filters are obtained imposing locality.

\subsection{Pooling}
The conventional spatial max pooling used in CNNs has two drawbacks in Spherical CNNs:
(1) need an expensive ISFT to convert back to spatial domain, and
(2) equivariance is not completely preserved, specially because of unequal cell areas from equiangular sampling.
Weighted average pooling (WAP) takes into account the cell areas to mitigate the latter, but is still affected by the former.

We introduce the spectral pooling (SP) for Spherical CNNs.
If the input has bandwidth $b$, we remove all coefficients with degree larger or equal than $b/2$ (effectively, a lowpass box filter).
Such operation is known to cause ringing artifacts, which can be mitigated by previous smoothing, although we did not find any performance advantage in doing so.
Note that spectral pooling was proposed before for conventional CNNs \cite{rippel15_spect_repres_convol_neural_networ}.

We found that spectral pooling is significantly faster, reduces the equivariance error, but also reduces classification accuracy.
The choice between SP and WAP is application-dependent.
For example, our experiments show SP is more suitable for shape alignment, while WAP is better for classification and retrieval.
Table~\ref{tab:ablation} shows the performance for each method.

\subsection{Global pooling}
\begin{figure}
\centering
\includegraphics[width=0.7\linewidth]{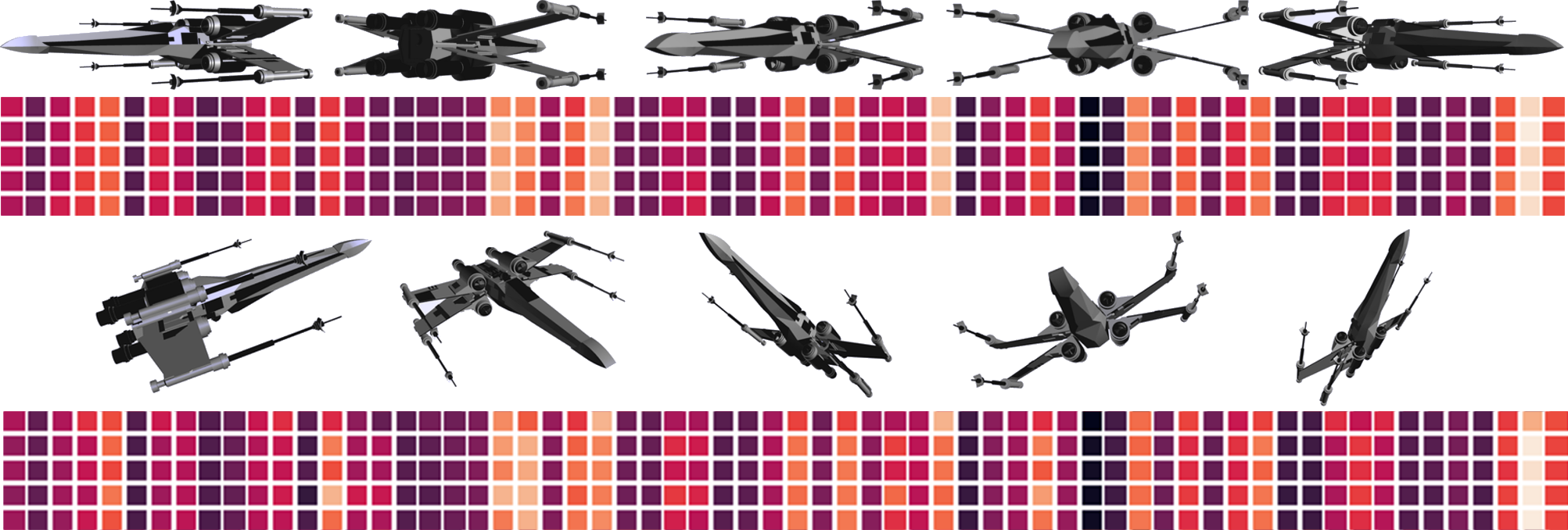}
\caption{
  Our model learns descriptors that are nearly invariant to input rotations.
  From top to bottom: azimutal rotations and correspondent descriptors (one per row), arbitrary rotations and correspondent descriptors.
  The invariance error is negligible for azimuthal rotations; since we use equiangular sampling, the cell area varies with the latitude, and rotations around $z$ preserve latitude.
  Arbitrary rotations brings a small invariance error, for reasons detailed in \ref{sec:equivariance}.
  }
\label{fig:invariance}
\end{figure}

In fully convolutional networks, it is usual to apply a global average pooling at the last layer to obtain a descriptor vector, where each entry is the average of one feature map.
We use the same idea; however, the equiangular spherical sampling results in cells of different areas, so we compute a weighted average instead, where a cell's weight is the sine of its latitude.
We denote it Weighted Global Average Pooling (WGAP).
Note that the WGAP is invariant to rotation, therefore the descriptor is also invariant.  Figure~\ref{fig:invariance} shows such descriptors.

An alternative to this approach is to use the magnitude per degree of the SFT coefficients; formally, if the last layer has bandwidth $b$ and
$\hat{f^{\ell}} = [\hat{f}_{-\ell}^{\ell},\hat{f}_{-\ell+1}^{\ell}, \ldots, \hat{f}_{\ell}^{\ell}]$, then
$d = \left[\norm{\hat{f}^0}, \norm{\hat{f}^1}, \ldots \norm{\hat{f}^{b-1}}\right]$
is an invariant descriptor \cite{arfken1966mathematical}.
We denote this approach as MAG-L (magnitude per degree $\ell$).
We found no difference in classification performance when using it (see Table~\ref{tab:ablation}).

\subsection{Architecture}

Our main architecture has two branches, one for distances and one for surface normals.
This performs better than having two input channels and slightly better than having two separate voting networks for distance and normals.
Each branch has 8 spherical convolutional layers, and $16,16,32,32,64,64,128,128$ channels per layer.
Pooling and feature concatenation of one branch into the other is performed when the number of channels increase.
WGAP is performed after the last layer, which is then projected into the number of classes.

%% file: experiments.tex
\section{Experiments}
\label{sec:experiments}

The greatest advantage of our model is inherent equivariance to \SO(3); we focus the experiments in problems that benefit from it;
namely, shape classification and retrieval in arbitrary orientations, and shape alignment.

We chose problems related to 3D shapes due to the availability of large datasets and published results on them;
our method would also be applicable to any kind of data that can be mapped to the sphere (e.g. panoramas).

\subsection{Preliminaries}

\subsubsection{Ray-mesh intersection:}
\label{sec:spherical-3d-object}
3D shapes are usually represented by mesh or voxel grid, which need to be converted to spherical functions.
Note that the conversion function itself must be equivariant to rotations; our learned representation will not be equivariant if the input is pre-processed by a non-equivariant function.

Given a mesh or voxel grid, we first find the bounding sphere and its center.
Given a desired resolution $n$, we cast $n \times n$ equiangular rays from the center, and obtain the intersections between each ray and the mesh/voxel grid.
Let $d_{jk}$ be the distance from the center to the farthest point of intersection, for a ray at direction $(\theta_j, \phi_k)$.
The function on the sphere is given by $f(\theta_j, \phi_k) = d_{jk},\, 1 \le j,k \le n$.

For mesh inputs, we also compute the angle $\alpha$ between the ray and the surface normal at the intersecting face, giving a second channel $f(\theta_j, \phi_k) =  [d, \sin\alpha]$.

Note that this representation is suitable for star-shaped objects, defined as objects that contain an interior point from where the whole boundary is visible.
Moreover, the center of the bounding sphere must be one of such points.
In practice, we do not check if these conditions hold -- even if the representation is ambiguous or non-invertible, it is still useful.

\subsubsection{Training:}
We train using ADAM, for 48 epochs, initial learning rate of $10^{-3}$, which is divided by 5 on epochs 32 and 40.

We make use of data augmentation for training, performing rotations, anisotropic scaling and mirroring on the meshes, and adding jitter to the bounding sphere center when constructing the spherical function.
Note that, even though our learned representation is equivariant to rotations, augmenting the inputs with rotations is still beneficial due to interpolation and sampling effects.

\subsection{3D object classification}

This section shows classification performance on ModelNet40 \cite{Wu20153Dshapenets}.
Three modes are considered:
(1) trained and tested with azimuthal rotations (z/z),
(2) trained and tested with arbitrary rotations (\SO(3)/\SO(3)),
and (3) trained with azimuthal and tested with arbitrary rotations (z/\SO(3)).

Table \ref{tab:m40} shows the results.
All competing methods suffer a sharp drop in performance when arbitrary rotations are present, even if they are seen during training.
Our model is more robust, but there is a noticeable drop for mode 3, attributed to sampling effects.
Since we use equiangular sampling, the cell area varies with latitude.
Rotations around $z$ preserve latitude, so regions at same height are sampled at same resolution during  training, but not during test.
We believe this can be improved by using equal-area spherical sampling.

We evaluate competing methods using default settings of their published code.
The volumetric \cite{Qi2016volumetric} and point cloud based \cite{qi2017pointnet,qi2017pointnet++}  methods cannot generalize to unseen orientations (z/\SO(3)).
The multi-view \cite{su2015multi,kanezaki16_rotat} methods can be seen as a brute force approach to equivariance; and MVCNN \cite{su2015multi} generalizes to unseen orientations up to a point.
Yet, the Spherical CNN outperforms it, even with orders of magnitude fewer parameters and faster training.
Interestingly, RotationNet \cite{kanezaki16_rotat}, which holds the current state-of-the-art on ModelNet40 classification, fails to generalize to unseen rotations, despite being multi-view based.

Equivariance to \SO(3) is unneeded when only azimuthal rotations are present (z/z); the full potential of our model is not exercised in this case.

\begin{table}
  \caption{ModelNet40 classification accuracy per instance.
    Spherical CNNs are robust to arbitrary rotations, even when not seen during training,
    while also having one order of magnitude fewer parameters and faster training. \label{tab:m40}
  }
  \centering
    {\def\arraystretch{1}\setlength\tabcolsep{5pt}
    \begin{tabular}{lcccrrr}
      \hline
      Method & z/z & SO3/SO3 & z/SO3 & params & inp. size \\
      \hline
      PointNet \cite{qi2017pointnet} & 89.2 & 83.6 & 14.7 & 3.5M & 2048 x 3 \\
      PointNet++ \cite{qi2017pointnet++} & 89.3 & 85.0 & 28.6 & 1.7M & 1024 x 3 \\
      VoxNet \cite{daniel2015voxnet} & 83.0 & 73.0 & - & 0.9M & $30^3$ \\
      SubVolSup \cite{Qi2016volumetric} & 88.5 & 82.7 & 36.6 & 17M & $30^3$ \\
      SubVolSup MO \cite{Qi2016volumetric} & 89.5 & 85.0 & 45.5 & 17M & $20 \times 30^3$ \\
      MVCNN 12x \cite{su2015multi} & 89.5 & 77.6 & 70.1 & 99M & $12 \times 224^2$ \\
      MVCNN 80x \cite{su2015multi} & \textbf{90.2} & 86.0 &- \footnotemark & 99M & $80 \times 224^2$ \\
      RotationNet 20x \cite{kanezaki16_rotat} & \textbf{92.4} & 80.0 & 20.2 & 58.9M & $20 \times 224^2$ \\
      Ours & 88.9 & \textbf{86.9} & \textbf{78.6} & \textbf{0.5M} & $\mathbf{2 \times 64^2}$ \\
      \hline
    \end{tabular}
    }
\end{table}
\footnotetext{The 80 views are not restricted to azimuthal, hence cannot be compared (acc: 81.5\%).}

\subsection{3D object retrieval}

We run retrieval experiments on ShapeNet Core55 \cite{chang15_shapen},
following the SHREC'17 3D shape retrieval rules \cite{savva2017shrec},
which includes random \SO(3) perturbations.

The network is trained for classification on the 55 core classes (we do not use the subclasses), with an extra in-batch triplet loss (from \cite{schroff2015facenet}) to encourage descriptors to be close for matching categories and far for non-matching.

The invariant descriptor is used with a cosine distance for retrieval.
We first compute a threshold per class that maximizes the training set F-score.
For test set retrieval, we return elements whose distances are below their class threshold and include all elements classified as the same class as the query.
Table~\ref{tab:shrec} shows the results.
Our model matches the state of the art performance (from \cite{furuya2016deep}), with significantly fewer parameters, smaller input size, and no pre-training.

\begin{table}
  \caption{SHREC'17 perturbed dataset results.
  We show precision, recall and mean average precision.
  \emph{micro} average is adjusted by category size, \emph{macro} is not.
  The sum of \emph{micro} and \emph{macro} mAP is used for ranking.
  We match the state of the art even with significantly fewer parameters, smaller input resolution, and no pre-training.
  Top results are bold, runner-ups italic.
}%
  \centering
  {\scriptsize\def\arraystretch{1}\setlength\tabcolsep{5pt}
    \newcommand{\ba}[1]{\textbf{#1}}
    \newcommand{\bb}[1]{\emph{#1}}

  \begin{tabular}{l|rrr|rrr|rrr}
    \hline
                                    & \multicolumn{3}{|c|}{micro} & \multicolumn{3}{c|}{macro} & total &                                \\
                                    & P@N                        & R@N        & mAP        & P@N        & R@N        & mAP        & score & input size               & params        \\
    \hline
    Furuya \cite{furuya2016deep}    & \ba{0.814}                 & 0.683      & 0.656      & \ba{0.607} & 0.539      & \ba{0.476} & \ba{1.13} & $126\times 10^3$         & 8.4M          \\
    Ours                            & \bb{0.717}                 & \bb{0.737} & \bb{0.685} & \bb{0.450} & \bb{0.550} & \bb{0.444} & \ba{1.13} & $\mathbf{2\times64^2}$   & \ba{0.5M}     \\
    Tatsuma \cite{tatsuma2009multi} & 0.705                      & \ba{0.769} & \ba{0.696} & 0.424      & \ba{0.563} & 0.418      & \bb{1.11} & $38\times224^2$          & 3M            \\
    Cohen \cite{s.2018spherical}    & 0.701                      & 0.711      & 0.676      & -          & -          & -          & - & $\mathit{6\times 128^2}$ & \textit{1.4M} \\
    Zhou \cite{bai2016gift}         & 0.660                      & 0.650      & 0.567      & 0.443      & 0.508      & 0.406      & 0.97 & $50\times224^2$          & 36M           \\
    \hline
  \end{tabular}
  }
\label{tab:shrec}
\end{table}

\subsection{Shape alignment}

Our learned equivariant feature maps can be used for shape alignment using spherical correlation.
Given two shapes from the same category (not necessarily the same instance), under arbitrary orientations, we run them through the network and collect the feature maps at some layer.
We compute the correlation between each pair of corresponding feature maps, and add the results.
The maximum value of the correlation function (which takes inputs on $\SO(3)$) corresponds to the rotation that aligns both shapes \cite{makadia2010spherical}.

Features from deeper layers are richer and carry semantic value, but are at lower resolution.
We run an experiment to determine the performance of the shape alignment per layer,
while also comparing with the spherical correlation done at the network inputs (not learned).

\begin{wraptable}{r}{0.4\textwidth}\vspace{-20pt}
  \caption{\small Shape alignment median angular error in degrees.
    The intermediate learned features are best suitable for this task.}
  \label{tab:alignment}
  \centering%
    {\tiny\def\arraystretch{1}\setlength\tabcolsep{5pt}
    \begin{tabular}{lrrrr}
      \hline
      & bed & chair & sofa & toilet\\
      \hline
      input & 91.63          & 111.47         & 12.15          & 21.65\\
      conv2 & 85.64          & 21.10          & 14.47          & 14.95\\
      \textbf{conv4} & \textbf{12.73} & \textbf{14.63} & \textbf{10.03} & \textbf{11.03}\\
      conv6 & 16.70          & 18.92          & 15.83          & 17.62\\
      \hline
    \end{tabular}
    }
    \vspace{-20pt}
  \end{wraptable}

\begin{figure}
\centering
\includegraphics[width=\linewidth]{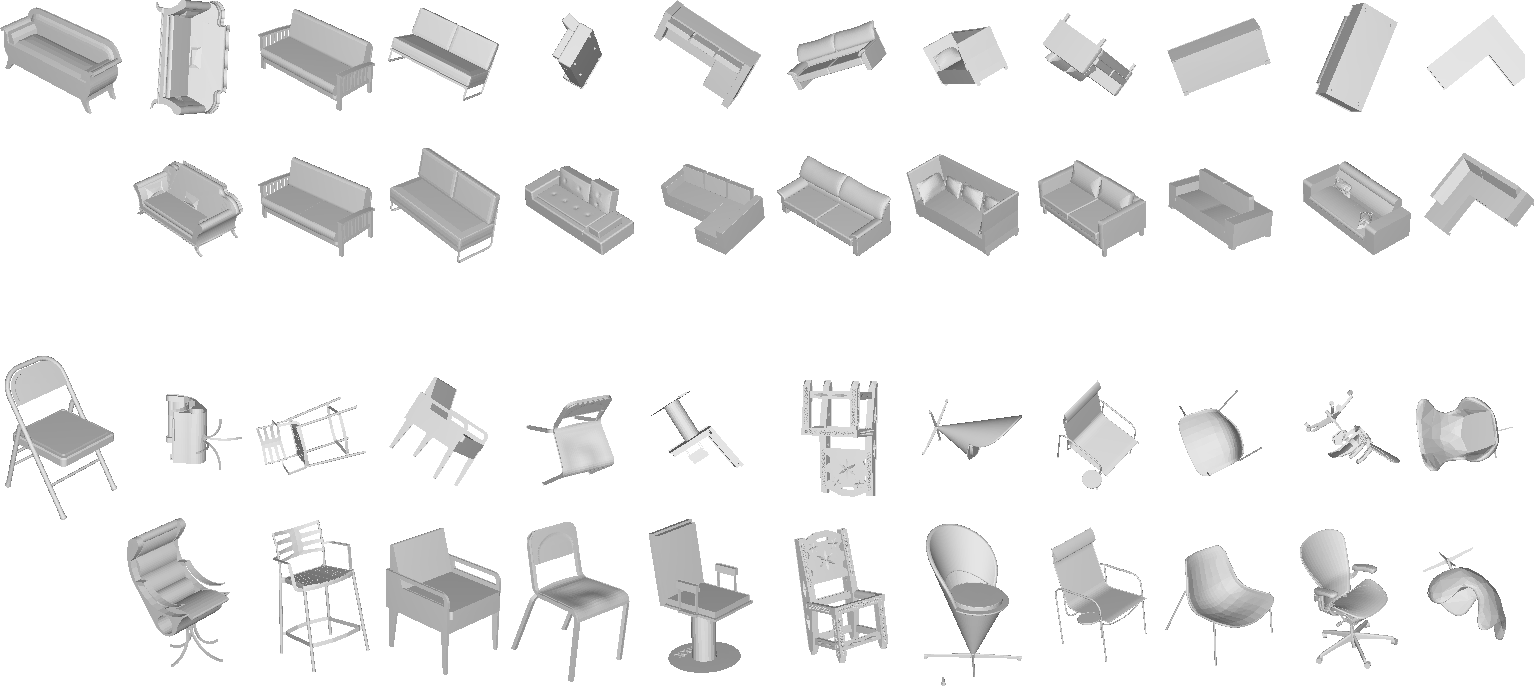}
\caption{
  Shape alignment for two categories.
  We align shapes by running spherical correlation of their feature maps.
  The semantic features learned can be used to align shapes from the same class even with large appearance variation.
  \emph{1st and 3rd rows:} reference shape, followed by queries from the same category.
  \emph{2nd and 4th rows:} Corresponding aligned shapes.
  Last column shows failure cases.
}
\label{fig:alignment}
\end{figure}

We select categories from ModelNet10 that do not have rotational symmetry so that the ground truth rotation is unique and the angular error is measurable.
These categories are: \emph{bed, sofa, toilet, chair}.
Only entries from the test set are used.
Results are in Table \ref{tab:alignment},
while Figure~\ref{fig:alignment} shows some examples.
Results show that the learned features are superior to the handcrafted spherical shape representation for this task, and best performance is achieved by using intermediate layers.
The resolution at conv4 is $32 \times 32$, which corresponds to cell dimensions up to $11.25 \text{ deg}$, so we cannot expect errors much lower than this.

\subsection{Equivariance error analysis}
\label{sec:equivariance}
Even though spherical convolutions are equivariant to \SO(3) for bandlimited inputs, and spectral pooling preserves bandlimit, there are other factors that may introduce equivariance errors.
We quantify these effects in this section.

We feed each entry in the test set and one random rotation to the network, then apply the same rotation to the feature maps and measure the average relative error.
Table~\ref{tab:equivariance} shows the results.
The pointwise nonlinearity does not preserve bandlimit, and cause equivariance errors (rows 1, 4).
The mesh to sphere map is only approximately equivariant, which can be mitigated with larger input dimensions (\emph{input} column for rows 1, 5).
Error is smaller when the input is bandlimited (rows 1, 7).
Spectral pooling is exactly equivariant, while max-pooling introduces higher frequencies and has larger error than WAP (rows 1, 2, 3).
Error for an untrained model demonstrates that the equivariance is by design and not learned (row 6).
Note that the error is smaller because the learned filters are usually high-pass, which increase the pointwise relative error.
A linear model with bandlimited inputs has zero equivariance error, as expected (row 8).

Note that even conventional planar CNNs will exhibit a degree of translational equivariance error introduced by max pooling and discretization.
\begin{table}
  \caption{Equivariance error.
    Error is zero for bandlimited inputs and linear layers.
  }
  \label{tab:equivariance}
  \centering
  {\scriptsize
    \begin{tabular}{|l|ccccc|ccccccc|}
      \hline
                    & \multicolumn{5}{|c|}{configuration} & \multicolumn{7}{c|}{error per layer} \\
                    & res.                                & blim.                                & pool  & linear & trained & input & conv1 & conv2 & conv3 & conv4 & conv5 & conv6 \\
      \hline
      1. baseline      & 64\(^{\text{2}}\)                   & no                                   & WAP   & no     & yes     & 0.05  & 0.11  & 0.12  & 0.14  & 0.16  & 0.17  & 0.15  \\
      2. maxpool       & 64\(^{\text{2}}\)                   & no                                   & max   & no     & yes     & 0.05  & 0.11  & 0.12  & 0.14  & 0.18  & 0.19  & 0.15  \\
      3. specpool      & 64\(^{\text{2}}\)                   & no                                   & SP    & no     & yes     & 0.05  & 0.11  & 0.12  & 0.10  & 0.10  & 0.09  & 0.08  \\
      4. linear        & 64\(^{\text{2}}\)                   & no                                   & WAP   & yes    & yes     & 0.05  & 0.12  & 0.13  & 0.15  & 0.14  & 0.12  & 0.04  \\
      5. lowres        & 32\(^{\text{2}}\)                   & no                                   & WAP   & no     & yes     & 0.09  & 0.15  & 0.18  & 0.21  & 0.21  & 0.21  & 0.20  \\
      6. untrained     & 64\(^{\text{2}}\)                   & no                                   & WAP   & no     & no      & 0.05  & 0.09  & 0.07  & 0.07  & 0.11  & 0.07  & 0.04  \\
      7. blim          & 64\(^{\text{2}}\)                   & yes                                  & WAP   & no     & yes     & 0.00  & 0.10  & 0.11  & 0.11  & 0.15  & 0.14  & 0.04  \\
      8. blim/lin/sp & 64\(^{\text{2}}\)                   & yes                                  & SP    & yes    & yes     & 0.00  & 0.01  & 0.01  & 0.00  & 0.00  & 0.00  & 0.00  \\
      \hline
\end{tabular}
  }
\end{table}
\subsection{Ablation study}

In this section we evaluate numerous variations of our method to determine the sensitivity to design choices.
First, we are interested in assessing the effects from our contributions SP, WAP, WGAP, and localized filters.
Second, we are interested in understanding how the network size affects performance.
Results show that the use of WAP, WGAP, and localized filters significantly improve performance,
and also that further performance improvements can be achieved with larger networks.
In summary, factors that increase bandwidth (e.g. max-pooling) also increase equivariance error and
may reduce accuracy.
Global operations in early layers (e.g. non-local filters) escape the receptive field and reduce accuracy.

\begin{table}
  \caption{Ablation study.
    Spherical CNN accuracy on rotated ModelNet40.
    We compare various types of pooling, filter localization and network sizes.
  }
  \label{tab:ablation}
  \centering
    {\scriptsize\def\arraystretch{1.2}\setlength\tabcolsep{5pt}
    \begin{tabular}{cccccc|r}
      \hline
      inp. res. & pool & global pool & localized & params & details & acc. [\%]\\
      \hline
      $64\times 64$ & WAP & WGAP & yes & 0.49M & best & \textbf{86.9}\\
      $64\times 64$ & WAP & MAG-L & yes & 0.54M &  & 86.9 \\
      $64\times 64$ & SP & WGAP & yes & 0.49M &  & 85.8\\
      $64\times 64$ & max & WGAP & yes & 0.49M &  & 86.7\\
      $64\times 64$ & avg & WGAP & yes & 0.49M &  & 86.7\\
      $64\times 64$ & WAP & avg & yes & 0.49M &  & 86.4\\
      $64\times 64$ & WAP & WGAP & no & 0.49M &  & 85.9\\
      \hline
      $32\times 32$ & WAP & WGAP & yes & 0.39M &  & 85.0 \\
      $32\times 32$ & WAP & WGAP & yes & 0.69M & deeper & 85.6\\
      $32\times 32$ & WAP & WGAP & yes & 1.06M & wider & 85.5 \\
      $32\times 32$ & WAP & WGAP & yes & 0.12M & narrower & 83.8\\
      \hline
    \end{tabular}
    }
\end{table}